\documentclass[lettersize,journal]{IEEEtran}
\usepackage{amsmath,amsfonts}
\usepackage{algorithmic}
\usepackage{array}
\usepackage[caption=false,font=normalsize,labelfont=sf,textfont=sf]{subfig}
\usepackage{textcomp}
\usepackage{stfloats}
\usepackage{url}
\usepackage{verbatim}
\usepackage{graphicx}
\def\BibTeX{{\rm B\kern-.05em{\sc i\kern-.025em b}\kern-.08em
    T\kern-.1667em\lower.7ex\hbox{E}\kern-.125emX}}
\usepackage{balance}
\usepackage{booktabs}
\usepackage{multirow}

\begin{document}

\title{Compositional Chain-of-Relations for Faithful Knowledge Graph Question Answering with Large Language Models}

\author{Chenhui Liu, Jianpeng Zhou, and Jiahai Wang,~\IEEEmembership{Senior Member,~IEEE}
\thanks{All authors are with Sun Yat-sen University, Guangzhou 510006, China. (Corresponding author: Jiahai Wang. E-mail: wangjiah@mail.sysu.edu.cn)}
\thanks{This work is supported by the National Natural Science Foundation of China (62472461), and the Guangdong Basic and Applied Basic Research Foundation (2025A1515010129).}}


\markboth{Journal of \LaTeX\ Class Files,~Vol.~18, No.~9, September~2020}%
{How to Use the IEEEtran \LaTeX \ Templates}

\maketitle

\begin{abstract}
Knowledge graph question answering (KGQA) is a key task for evaluating KG-augmented Large Language Models (LLMs), and complex KGQA that requires multi-hop reasoning is especially challenging.
Solving a complex query involves two coupled phases: candidate retrieval, which locates answer candidates over the KG, and constraint handling, which filters these candidates against the query constraints.
Faithful reasoning requires grounding both phases in the KG.
However, existing agent-based methods ground candidate retrieval through entity-centric exploration, while leaving constraint handling to the LLM's internal knowledge, which leads to two critical limitations.
(1)~Unreliable entity pruning: entity-centric exploration uses entities as search units and must prune them to a fixed-size subset at each hop. Because entity information in KGs is often incomplete and a fixed-size subset cannot retain all valid entities, such pruning inevitably drops valid entities and ultimately leads to wrong answers.
(2)~Ungrounded constraint handling: query constraints are resolved from the LLM's internal knowledge rather than the KG, leaving the final answers unverifiable and prone to hallucination.
To address these limitations, this paper introduces a \textbf{relation-centric exploration paradigm}, which uses relations rather than entities as search units and thus avoids unreliable entity pruning.
Built on this paradigm, this paper proposes \textbf{Compositional Chain-of-Relations (CCoR)}, a simple and effective framework that grounds both phases in the KG with two relation chains: a main chain for candidate retrieval and a constraint chain that verifies query constraints through explicit KG exploration.
Experiments on four KGQA benchmarks show that CCoR consistently improves accuracy, faithfulness, and efficiency over strong baselines, with more pronounced gains on complex queries.
\end{abstract}


\section{Introduction}

Large Language Models (LLMs) have demonstrated remarkable reasoning capabilities across various tasks~\cite{zhao2024explainability}, particularly in natural language understanding and generation.
They also encode broad world knowledge from large corpora and adapt well to diverse tasks through prompting and instruction tuning.
However, their knowledge remains implicitly stored in parameters, making factual verification difficult and leaving persistent issues of outdated knowledge~\cite{fan2024survey}, hallucinations~\cite{huang2025survey, lecu2025reducing}, and limited explainability~\cite{zhao2024explainability}.
Knowledge graphs (KGs)~\cite{ji2021survey}, by contrast, represent facts explicitly as entity-relation triples and support queryable, verifiable, and updatable knowledge with transparent and controllable reasoning paths. These properties make KGs a natural complement to LLMs and create new opportunities for building faithful AI systems~\cite{pan2024unifying, delong2025neurosymbolic, jin2024large}.

\begin{figure}[t]
\centering
\includegraphics[width=0.98\columnwidth]{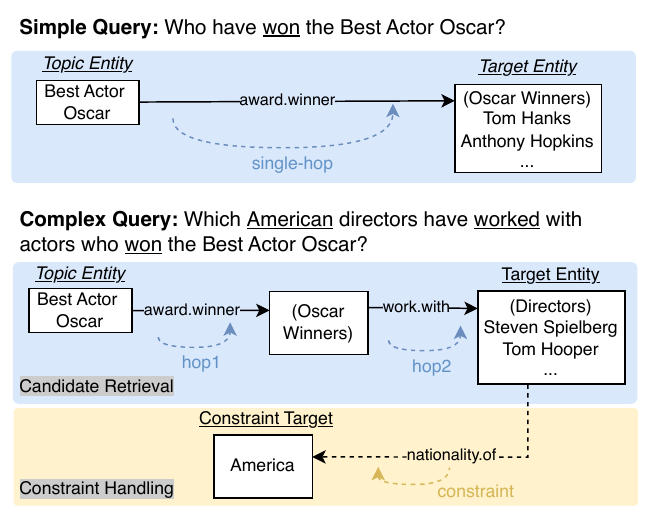}
\caption{Illustration of simple and complex KGQA queries and the two phases involved in solving complex queries. The simple query, ``Who has won the Best Actor Oscar?'', is answerable via one-hop reasoning (\texttt{award.winner}). The complex query, ``Which American directors have worked with actors who won the Best Actor Oscar?'', requires multi-hop reasoning and a constraint, and can be decomposed into two phases. \textit{Candidate retrieval} applies \texttt{award.winner} $\rightarrow$ \texttt{work.with} from the topic entities to retrieve candidate directors, while \textit{constraint handling} applies \texttt{nationality.of} with constraint target ``America'' to filter these candidates into the final answers.}
\label{fig:intro1}
\end{figure}

Knowledge graph question answering (KGQA) aims to answer natural language questions by retrieving facts from knowledge graphs~\cite{talmor2018web, yih2016value}.
This task unifies natural language understanding and KG reasoning in a single task, making it a key benchmark for KG-enhanced LLMs.

\begin{figure*}[htbp]
\centering
\includegraphics[width=0.98\textwidth]{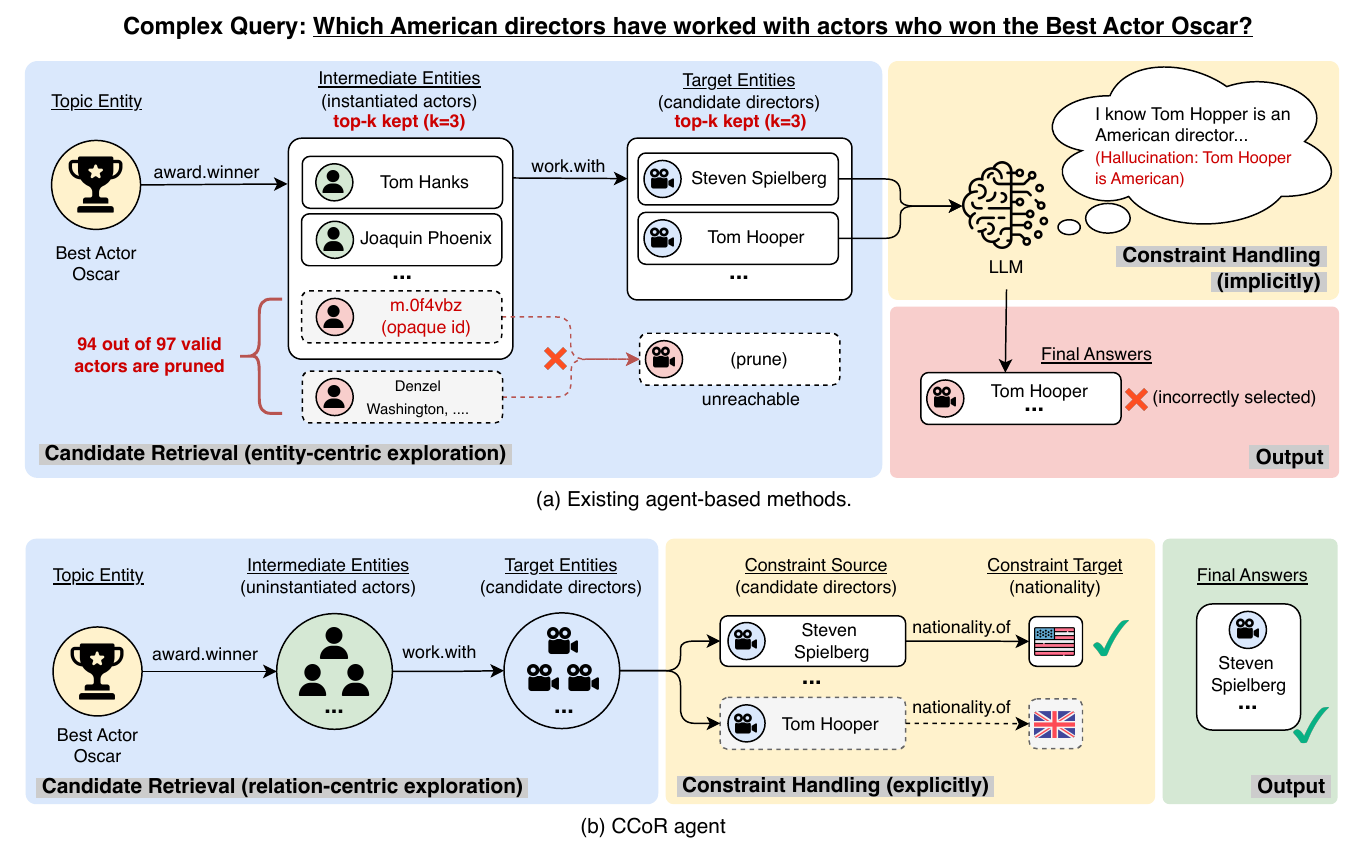}
\caption{Comparison between existing agent-based KGQA methods and CCoR on the complex query ``Which \textit{American} directors have worked with actors who won the Best Actor Oscar?''. (a)~Agent-based methods suffer from two limitations: (i)~they rely on unreliable entity pruning that frequently discards valid reasoning paths, and (ii)~they handle constraints implicitly through the LLM's parametric knowledge, making them prone to factual hallucinations. (b)~CCoR grounds both candidate retrieval and constraint handling in KG-executable relation chains, thereby producing faithful answers.}
\label{fig:intro2}
\end{figure*}

In particular, complex KGQA~\cite{lan2022complex} poses a more demanding challenge because it requires multi-hop reasoning and constraint handling over KGs. 
As illustrated in Figure~\ref{fig:intro1}, this work distinguishes two query types: a simple query, answerable through single-hop KG reasoning without constraints~\cite{hu2023empirical}, and a complex query, which requires either multi-hop KG reasoning or constraint handling, or both in a compositional manner. 
Accordingly, solving complex queries involves two coupled phases: (1) \textbf{candidate retrieval} starts from topic entities and retrieves target entities in the KG as answer candidates for the query, while (2) \textbf{constraint handling} filters these candidates using constraints in the query.

Existing methods devote mostly their design to candidate retrieval, casting it as the central problem. Along this line, prior work has integrated KGs into LLM reasoning through pre-training~\cite{zhang2019ernie, wang2021kepler, qiu2024knowledge}, fine-tuning~\cite{zhang2023fc, gu2023don}, or subgraph retrieval~\cite{wang2024boosting, liao2026enhancing}. 
These approaches often rely on static knowledge injection or fixed retrieved context, limiting active exploration for complex multi-hop reasoning. More recently, agent-based approaches~\cite{jiang2023structgpt, sunthink, chen2024plan} allow iterative exploration, but most still follow an \textbf{entity-centric exploration} paradigm dominated by entity pruning. 
Under this paradigm, entities serve as the primary search units, and at each step only a small subset of entities is retained for further expansion.
The problem is that these pruning decisions are inherently unreliable: they must be made before sufficient entity information is collected, yet such information is frequently missing~\cite{sun2023multi}, so valid entities are often discarded, interrupting the reasoning path and causing recall loss. 
This paper refers to this critical limitation as \textbf{unreliable entity pruning}, and it arises for two distinct reasons.
First, the pruning decision may be infeasible because many KG entities lack a readable literal name and appear only as opaque IDs (e.g., \texttt{m.0f4vbz} without ``Adrien Brody''). 
As shown in Figure~\ref{fig:intro2}(a), such entities cannot be evaluated and may be skipped, thereby discarding directors who collaborated with this Oscar-winning actor.
Second, even when entities are readable, the pruning decision may be premature. To control the combinatorial growth of exploration, entity-centric methods often perform top-$k$ pruning before sufficient KG evidence is collected. As illustrated in Figure~\ref{fig:intro2}(a), pruning 97 Best Actor Oscar winners to only 3 under a beam width of $k{=}3$ discards 94 actors, making directors connected exclusively to them unreachable.

Constraint handling, by contrast, has received far less attention. Earlier work addresses constraints through semantic parsing~\cite{bao2016constraint, chen2022outlining, qi2024enhancing} or embedding-based filtering~\cite{saxena2020improving, he2021improving, yasunaga2021qa}, translating constraints into executable logical forms or scoring candidates in a learned vector space, yet these approaches rely on predefined schemas or fixed representations, limiting flexible constraint resolution for diverse complex queries. 
Recently, agent-based approaches~\cite{jiang2023structgpt, sunthink, chen2024plan} take a shortcut that offloads constraint handling onto the LLM, resolving constraints implicitly from its internal knowledge instead of exploring on the KG~\cite{liu-etal-2026-chain}. 
As a consequence, constraints are never verified against the KG, resulting in unfaithful reasoning and hallucination, precisely the weakness that KG-augmented LLM reasoning is intended to prevent. This paper refers to this critical limitation as \textbf{ungrounded constraint handling}. 
As shown in Figure~\ref{fig:intro2}(a), when answering ``Which \textit{American} directors have worked with actors who won the Best Actor Oscar?'', the LLM may rely on parametric memory and hallucinate that director Tom Hooper is American, while he is actually British. Because this nationality judgment is made without KG verification, the method retains a wrong candidate and produces incorrect final answers.

This paper addresses the two limitations progressively.
To address \textbf{unreliable entity pruning}, this work introduces the \textbf{relation-centric exploration} paradigm and proposes the \textbf{Chain-of-Relations (CoR)} framework~\cite{liu-etal-2026-chain}, which uses relations rather than entities as the search units. 
Since intermediate entities are no longer evaluated or pruned during exploration, CoR structurally avoids unreliable entity pruning and casts candidate retrieval as a \textit{main chain}, a relation chain from topic entities to target entities that reaches a candidate answer space without intermediate entity pruning. 
As illustrated in Figure~\ref{fig:intro1}(b), for the query ``Which American directors have worked with actors who won the Best Actor Oscar?'', the main chain starts from the topic entity ``Best Actor Oscar'' and follows \texttt{award.winner} and \texttt{work.with} to construct the candidate space of directors (e.g., Steven Spielberg and Tom Hooper).
To address \textbf{ungrounded constraint handling}, this work further extends CoR into \textbf{Compositional Chain-of-Relations (CCoR)} by introducing a \textit{constraint chain}, a second relation chain from a constraint source to a constraint target that grounds constraint handling in explicit KG exploration. 
Continuing the example in Figure~\ref{fig:intro1}(b), the constraint chain applies \texttt{nationality.of} on the constraint source (candidate directors) with the constraint target ``America'', filtering out non-American candidates to produce the final answers. 
The constraint chain is constructed on demand and activated only when an explicit constraint is identified in a complex query. 
By composing the main chain with constraint chain, CCoR grounds both candidate retrieval and constraint handling in explicit KG exploration rather than parametric assumptions, yielding a more faithful and verifiable reasoning process.

Additionally, existing answer correctness metrics (e.g., Hits@1, F1) are insufficient to characterize model faithfulness. 
A method may produce correct answers by relying on parametric memory recall rather than explicit KG exploration, resulting in an unverifiable reasoning process.
To bridge in this gap, this paper introduces two faithfulness metrics that distinguish answers derived from these two pathways.
Specifically, \textbf{Knowledge-Grounded Rate (KGR)} measures how often answers are grounded in explicit KG exploration, and \textbf{Faithful-F1 (FF1)} jointly captures answer accuracy and grounded reasoning quality.

This work is an extended version of conference paper CoR~\cite{liu-etal-2026-chain}.
Compared with CoR, this extension introduces substantial improvements in three aspects.
(1)~At the method level, CCoR targets the ungrounded constraint handling limitation, introduces constraint-chain exploration, and composes it with main-chain exploration, enabling fully grounded reasoning via explicit KG exploration.
(2)~At the evaluation level, KGR is redefined from a binary indicator into a fine-grained weighted grounding score that distinguishes three levels of grounding, namely answers derived fully from KG, partially from KG, or entirely from LLM parametric memory. Building on this refined KGR, this paper introduces Faithful-F1 to unify answer accuracy and reasoning faithfulness in a single metric.
(3)~At the experimental level, additional backbone models, including GPT-3.5-turbo and GPT-5.4-mini, are evaluated with enriched baseline comparisons, wider benchmark coverage, and more comprehensive analyses across complex-query settings.

The main contributions of this work are as follows:
\begin{itemize}
	\item \textbf{Relation-centric exploration paradigm.} This work introduces the relation-centric exploration paradigm, which uses relations rather than entities as the search units, avoiding unreliable entity pruning during candidate retrieval over KGs.
	\item \textbf{CCoR framework.} Building on the relation-centric exploration paradigm, this work proposes CCoR, a unified framework that mirrors candidate retrieval and constraint handling with two compositional relation chains. By executing both chains through KG exploration, CCoR grounds both phases in the KG and thereby improves reasoning faithfulness.
	\item \textbf{Faithfulness evaluation.} This work introduces KGR to measure how strongly answers are grounded in KG and FF1 to jointly evaluate answer correctness and reasoning faithfulness.
	\item \textbf{Comprehensive experiments.} Experiments on four benchmarks (WebQSP, CWQ, QALD10, and GrailQA) with multiple LLM backbones demonstrate that CCoR consistently improves accuracy, faithfulness, and efficiency over strong baselines.
\end{itemize}

\section{Related Work}

\subsection{LLM Reasoning}

General LLM reasoning methods~\cite{xu2025large} rely on parametric knowledge learned during pre-training~\cite{hu2023survey, zhao2025survey}. Chain-of-Thought (CoT)~\cite{wei2022chain} prompting~\cite{wei2022chain} encourages LLMs to generate step-by-step rationales before producing final answers, Tree-of-Thought (ToT)~\cite{yao2023tree} organizes intermediate reasoning as tree-structured search, and Graph-of-Thought (GoT)~\cite{besta2024graph} extends this to graph-structured reasoning for richer dependency modeling. Self-reflective frameworks such as Reflexion~\cite{shinn2023reflexion} further introduce iterative verbal feedback for self-correction. However, these methods still operate largely within a closed language space without verified external symbolic evidence, so their reasoning remains dependent on parametric memory and is vulnerable to hallucination and outdated knowledge~\cite{huang2025survey, fan2024survey}. 
These limitations motivate KG reasoning~\cite{li2025knowledge, fan2024flow}, which anchors inference in explicit and verifiable KG facts. KGQA is a representative task of this paradigm, where simple queries are solved through candidate retrieval, while complex queries additionally require constraint handling.

\subsection{KG-Augmented LLM Reasoning}

Before the LLM era, traditional KGQA methods operated over predefined schemas or learned representations. 
Semantic-parsing methods~\cite{bao2016constraint, chen2022outlining, qi2024enhancing} translate a question into an executable logical form and run it over the KG, so that both the multi-hop relations and the constraints are enforced explicitly as logical-form operators. 
Embedding-based methods~\cite{saxena2020improving, he2021improving, yasunaga2021qa} instead score answer candidates in a learned vector space, while pre-training methods~\cite{zhang2019ernie, wang2021kepler, hu2023empirical} inject entity and relation facts into model parameters so that KG knowledge can be recalled from the pre-trained weights. These methods are effective on their target benchmarks, but their reliance on predefined schemas or static representations limits their flexibility on diverse complex queries.

The rise of LLMs~\cite{cheng2025survey} has reshaped KGQA into three lines of work that differ in how the LLM is coupled with the KG. 
\textit{Tuning-based} methods adapt the LLM on KG-derived supervision to internalize reasoning patterns over the target KG~\cite{zhang2023fc, gu2023don, li2024flexkbqa, shu2024distribution}. Aligning the model with the target schema lets it answer many queries directly, but constraints are resolved implicitly inside the fine-tuned parameters, and the required retraining binds each model to a specific KG and schema, leaving it costly to transfer.

\textit{Retrieval-based} methods~\cite{peng2025graph} keep the reasoning LLM frozen and feed it KG evidence as in-context prompts, such as subgraphs~\cite{baek2023knowledge, axelsson2023using, wang2024boosting}, or reasoning paths~\cite{luoreasoning, liao2026enhancing}. 
Here answer quality hinges on a separate retrieval module, typically a supervised-trained neural retriever that generalizes poorly once the KG or schema items shift. Moreover, the retrieved evidence only serves as a prompt and the final answer is still generated by the LLM, so the output remains susceptible to hallucination.

\textit{Agent-based} methods let the LLM traverse the KG iteratively and refine its exploration as new evidence is collected~\cite{jiang2025kg, liu2025symagent}. StructGPT~\cite{jiang2023structgpt} exposes structured interfaces to read and query KG facts step by step; ToG~\cite{sunthink, mathink} runs beam search over entity paths; and PoG~\cite{chen2024plan} plans the reasoning route in advance and revises it through self-correction. 
Despite interactive exploration, these methods still suffer from two limitations. First, they rely on unreliable entity pruning that discards valid intermediate entities during multi-hop traversal. Second, they resolve constraints with parametric knowledge rather than the KG, leaving the answers ungrounded and prone to hallucination.

In contrast to the entity-centric agent-based methods above, CoR~\cite{liu-etal-2026-chain} removes entity pruning from the exploration loop and instead uses relations as the basic search unit, building executable relation chains that avoid discarding valid intermediate entities and thus make KG exploration more reliable. Building upon relation-centric exploration paradigm, CCoR further models each constraint as a relation chain and grounds it over the KG, replacing parametric constraint judgment with explicit KG execution and thereby yielding faithful results.

\begin{table}[ht]
\centering
\caption{Notation Summary for CCoR}
\label{tab:notation_ccor}
\small
\setlength{\tabcolsep}{4pt}
\begin{tabular}{@{}l>{\raggedright\arraybackslash}p{0.66\columnwidth}@{}}
\toprule
Symbol & Description \\
\midrule
$G=(E,R,T)$ & Knowledge graph \\
$q$ & Natural language question \\
$E_{topic}$ & Topic entities in $q$ \\
$E_{cand}$ & Candidates from main-chain execution \\
$E_{con}$ & Feasible set from constraint-chain execution \\
$E_{final}$ & Final answer entity set \\
$c_{main}$ & Main chain for candidate construction \\
$c_{con}$ & Constraint chain \\
$\tau$ & Constraint target \\
$\kappa$ & Unresolved explicit constraint extracted from $q$ \\
$\mathcal{F}_{main}$ & Execution operator for main chain \\
$\mathcal{F}_{con}$ & Joint execution operator composing main and constraint chains \\
$s^t=(c^t,h^t)$ & Exploration state at step $t$ \\
$h^t$ & Memory stack for backtracking \\
$a_i$ & Final state label of sample $i$ \\
$w(a_i)$ & State weight for KGR \\
\bottomrule
\end{tabular}
\end{table}

\section{Preliminary}

This section introduces the key concepts and notation used throughout the CCoR framework.
Table~\ref{tab:notation_ccor} summarizes the main symbols used in this paper.

\noindent \textbf{Knowledge Graph.}
A knowledge graph is denoted as $G=(E,R,T)$, where $E$ is the entity set, $R$ is the relation set, and $T$ is the triple set. Each triple $(e_h,r,e_t)\in T$ encodes a factual relation between entities.

\noindent \textbf{Knowledge Graph Question Answering.}
Given a question $q$ and a knowledge graph $G$, KGQA aims to identify answer entities from $G$. Starting from topic entities $E_{topic}$ mentioned in $q$, CCoR performs relation-centric exploration to construct a candidate set, then applies explicit constraint execution to obtain the final answer set $E_{final}$.

\noindent \textbf{Main Chain and Constraint Chain.}
CCoR decomposes query solving into two complementary relation chains. The \textbf{main chain} $c_{main}$ connects topic entities to a candidate answer space, and its execution defines
\begin{equation}
\label{eq:main_exec}
E_{cand}=\mathcal{F}_{main}(G,E_{topic},c_{main}).
\end{equation}
For an unresolved explicit constraint $\kappa$ (if any), a \textbf{constraint chain} $c_{con}$ connects a constraint source to a constraint target $\tau$. Once both are determined, the constraint chain is composed with the main chain and jointly executed on $G$ to identify which candidates satisfy the constraint:
\begin{equation}
\label{eq:con_exec}
E_{con}=\mathcal{F}_{con}(G,c_{main},c_{con},\tau).
\end{equation}
The final answer set is obtained by compositional filtering when constraint-chain execution is triggered:
\begin{equation}
\label{eq:final_comp}
E_{final}=E_{cand}\cap E_{con}.
\end{equation}
When no explicit constraint is extracted, this reduces to $E_{final}=E_{cand}$. This decomposition separates candidate-space construction from constraint verification while preserving end-to-end compositional reasoning.

\begin{figure*}[htbp]
\centerline{\includegraphics[width=1.0\linewidth]{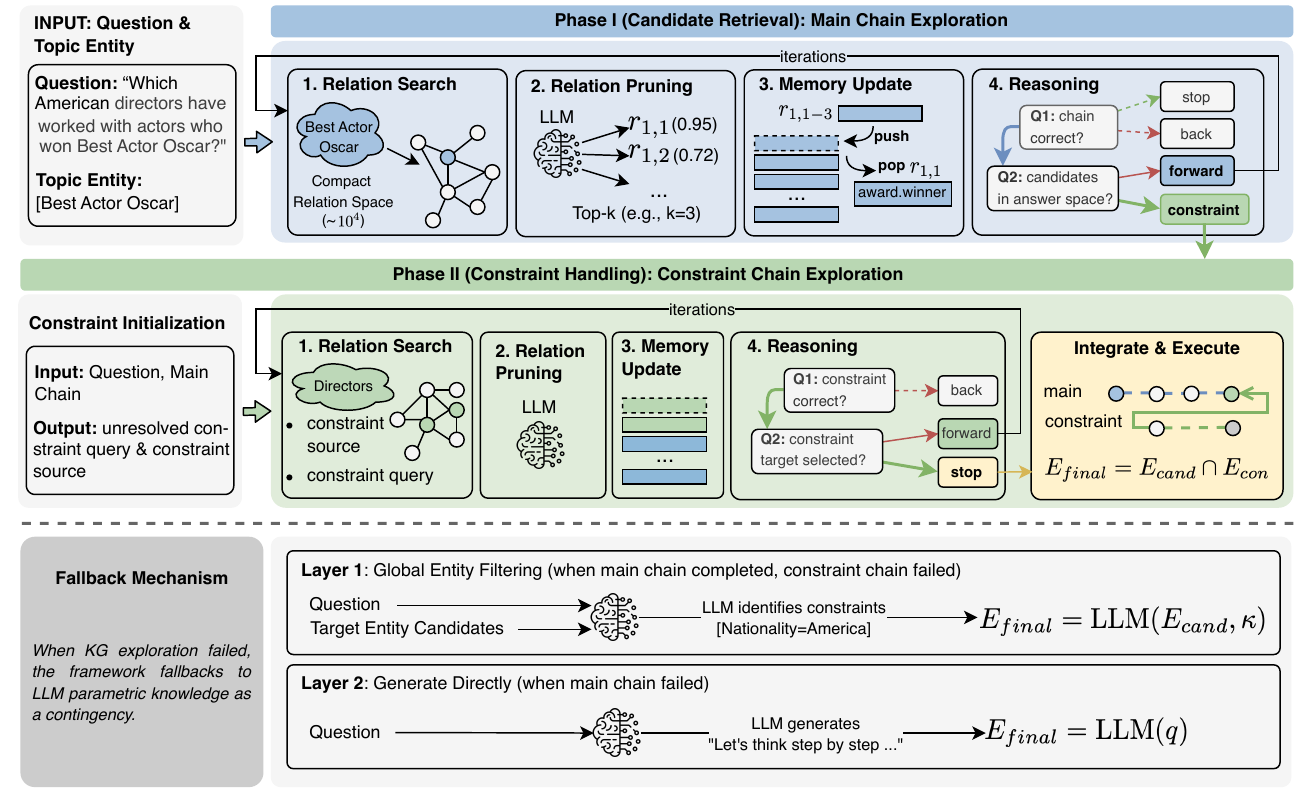}}
\caption{Architecture of the CCoR framework, illustrated with the question ``Which American directors have worked with actors who won the Best Actor Oscar?'' Main-Chain Exploration constructs candidate answers via relation-centric search from topic entities. Constraint-Chain Exploration then explicitly executes the nationality constraint through KG queries to filter candidates. A two-layer fallback mechanism handles failures: Layer~1 falls back to Global Entity Filtering when constraint-chain exploration fails, and Layer~2 falls back to parametric generation when main-chain reasoning fails.}
\label{fig:framework}
\end{figure*}

\section{CCoR Framework}

CCoR consists of two components and one auxiliary mechanism. The two components are Main-Chain Exploration for candidate construction and Constraint-Chain Exploration for explicit constraint execution. Both components are built on a shared relation-centric exploration engine that performs iterative chain construction in relation space. The auxiliary mechanism is a two-layer fallback strategy that preserves robustness when exploration fails. 
Through this design, CCoR converts constraint handling into auditable and executable KG operations, while maintaining reliable reasoning across queries of varying complexity.

\subsection{Relation-Centric Exploration Engine}

The relation-centric exploration engine is the shared inference foundation underlying both main-chain and constraint-chain exploration. At the beginning of exploration, the state is initialized as $s^0=(c^0,h^0)$, where $c^0$ is an empty chain and $h^0$ is an empty memory stack. At each step $t$, the state is updated to $s^t=(c^t,h^t)$, where $c^t$ is the current relation chain and $h^t$ stores explored branches and alternatives for backtracking.

Each iteration contains four steps: relation search, relation prune, memory update, and reasoning.

\subsubsection{Relation Search}
This step identifies candidate relations that can extend the current chain without explicit instantiation of intermediate entities. Given a relation chain $c^t=[r_1,\dots,r_t]$ from $E_{topic}$, the system queries the KG to retrieve expandable relation candidates $R_{cand}$. Similar to variable binding in SPARQL, intermediate entities are treated as implicit variables. The search is therefore conducted in compact relation space rather than large entity space.

\subsubsection{Relation Pruning}
This step scores each $r\in R_{cand}$ by its relevance to the question and consistency with the current chain. The score $score(r\mid q,c^t)$ is computed from three criteria: semantic alignment with question intent, logical consistency with chain context, and goal proximity to likely answer regions. The top-ranked relations are retained:
\begin{equation}
\label{eq:relation_prune_topk}
R_{prune}=\mathrm{Top}\text{-}k(R_{cand}, score).
\end{equation}

\subsubsection{Memory Update}
After pruning, candidate chains are formed by appending each retained relation to the current chain, i.e., $\{c^t \oplus r_i \mid r_i\in R_{prune}\}$. 
These candidates are pushed onto the memory stack in ascending order of their scores, yielding the updated memory $h^{t+1}$. As a result, the highest-scoring chain lies on the top of the stack and is popped first for the next expansion. 
This strategy preserves alternative branches while prioritizing promising chains, enabling systematic recovery through backtracking.

\subsubsection{Reasoning}
Given the current chain $c^t$ and the updated memory $h^{t+1}$, the agent reasons over the question context and chain progress to select the next action. The selected action determines the next chain $c^{t+1}$, and the next iteration begins from $s^{t+1}=(c^{t+1},h^{t+1})$. The available action set depends on the active component and is detailed in the corresponding subsection. Across both components, action selection adopts a hierarchical reasoning pattern that decomposes the action decision into two simple judgment questions. The combined answers to these questions determine the final action, reducing planning complexity in multi-hop exploration.

\subsection{Main Chain Exploration}

The main chain is a relation chain from $E_{topic}$ to target entities $E_{cand}$. This component applies the shared relation-centric exploration engine to construct a robust candidate space. Relation search, relation pruning, and memory update follow the shared engine with only minor prompt adaptations\footnote{See Appendix-A CCoR Prompt Templates}. The key specialization is in reasoning, where main-chain reasoning defines four actions:
\begin{itemize}
	\item \textit{Forward:} trigger a new exploration iteration, where relation search and pruning produce the next promising relation candidates.
	\item \textit{Backtrack:} discard the current branch when it is logically inconsistent and pop the next candidate relation from the memory stack to resume exploration from an alternative branch.
	\item \textit{Stop:} finalize the main chain when it is complete and no explicit constraint remains.
	\item \textit{Constraint:} finalize the main chain and hand off to constraint-chain exploration for explicit constraint execution.
\end{itemize}

Action selection follows a hierarchical reasoning process driven by two sequential questions:

\begin{itemize}
\item \textit{Is the current main chain logically correct and complete?} The agent evaluates whether the chain is consistent with the main reasoning path of the question up to the current step. Three outcomes are possible. (i)~The chain is correct and already complete, with no constraint remaining: the agent selects \textit{Stop}. (ii)~The chain is logically inconsistent: the agent selects \textit{Backtrack}. (iii)~The chain is correct so far but not yet complete: the process proceeds to the second question to determine the cause of incompleteness.
\item \textit{Have the current target entities reached the answer space of the question?} This question checks whether the type of the current target entities matches what the question asks for. For instance, if the question asks for a location, the current target entities should be places. If \textit{No}, the main reasoning path is still incomplete and the chain requires further extension, so the agent selects \textit{Forward}. If \textit{Yes}, the candidate space is established and the remaining incompleteness is due to an unhandled explicit constraint, so the agent selects \textit{Constraint} to begin explicit constraint execution.
\end{itemize}

Both \textit{Stop} and \textit{Constraint} finalize the main chain, outputting $c_{main}$. Upon finalization, the agent constructs a SPARQL query from the current main chain and executes it on the KG to obtain the corresponding candidate set $E_{cand}$.
Under \textit{Stop}, the final answer set is directly given as follows:
\begin{equation}
\label{eq:main_stop_final}
E_{final}=E_{cand}.
\end{equation}
Under \textit{Constraint}, $E_{cand}$ is passed to constraint-chain exploration for explicit filtering.

\subsection{Constraint Chain Exploration}

This component is triggered by the \textit{Constraint} action from main-chain reasoning. A constraint chain is a relation chain from a constraint source to a constraint target, and its purpose is to explicitly filter the current target entities. 
The component proceeds in two steps: \textit{Initialization} and \textit{Exploration}.

\subsubsection{Initialization}
In this step, given the question $q$ and the main chain, the agent outputs an unresolved constraint query and extracts a constraint source.
The unresolved constraint query reformulates the remaining constraint in the original question as a standalone sub-question, focusing on what the main chain has not yet addressed.
For example, from the original question ``Which American directors have worked with actors who won the Best Actor Oscar?'', the agent refines the unresolved constraint into ``Among all directors, who is from America?''
The constraint source is the entity to which this refined query applies, namely the directors already retrieved by the main chain. Formally, let $e_1,\dots,e_n$ denote the entities along the main chain, and the agent selects the one that serves as the subject of the unresolved constraint. 
In this example, the main chain produces the Oscar-winning actors $e_1$ and then the directors $e_2$, where $e_2$ is also the target entity. The subject of ``who is from America?'' is the directors $e_2$ rather than the intermediate actors $e_1$, so $e_2 (\texttt{target\_entity})$ is selected as the constraint source.

\subsubsection{Exploration}
In this step, constraint-chain exploration reuses the shared relation-centric exploration engine to construct $c_{con}$. Relation search, relation pruning, and memory update follow the shared engine. The key specialization is in reasoning, where constraint-chain reasoning defines three actions:
\begin{itemize}
	\item \textit{Forward:} trigger a new exploration iteration, where relation search and pruning produce the next promising relation candidates.
	\item \textit{Backtrack:} discard the current branch when it is invalid and pop the next candidate relation from the memory stack to resume exploration from an alternative branch.
	\item \textit{Stop:} execute the current chain and return filtered results for the current constraint.
\end{itemize}

Action selection follows a hierarchical reasoning process driven by two questions:

\begin{itemize}
	\item \textit{Is the current constraint chain logically correct for the unresolved constraint query?} The agent evaluates whether the chain is consistent with the unresolved constraint query up to the current step. If \textit{No}, the chain is logically inconsistent and the agent selects \textit{Backtrack} to resume from an alternative branch. If \textit{Yes}, the process proceeds to the second question.
	\item \textit{Can a valid constraint target be selected from the current options?} This question checks whether the current constraint chain has recalled candidates from which a concrete constraint target can be identified. If \textit{Yes}, a valid constraint target is available and the agent selects \textit{Stop} to output it. If \textit{No}, the constraint chain has not yet reached the constraint target and requires further extension, so the agent selects \textit{Forward} to continue constraint-chain expansion.
\end{itemize}

When \textit{Stop} is selected for the unresolved constraint $\kappa$, the agent outputs a concrete constraint target $\tau$ and constructs a single SPARQL query from both the main chain and the constraint chain, which is executed on $G$ to compute the feasible set $E_{con}$. After constraint handling, the final answer set is obtained by explicit filtering:
\begin{equation}
\label{eq:final_after_constraints}
E_{final}=E_{cand}\cap E_{con}.
\end{equation}

This component directly addresses ungrounded constraints by converting constraint handling from implicit memory-based judgment to explicit KG execution.

\subsection{Fallback Mechanism}

Exploration over KGs may fail in complex query settings (e.g., chain-construction failure or unreachable paths). To improve robustness, CCoR introduces a two-layer fallback mechanism as an auxiliary strategy. When explicit KG execution cannot be completed, the framework backs off to LLM parametric knowledge as a contingency. Each layer is triggered by a specific failure cause and provides a different level of KG grounding for the final answer.

\subsubsection{Layer 1: Global Entity Filtering}

If the main chain has been successfully constructed but constraint-chain exploration fails, the system falls back to Global Entity Filtering (GEF) for implicit constraint handling. After the main chain is executed to obtain $E_{cand}$, GEF uses the LLM to identify the unresolved constraint $\kappa$ directly from the question $q$ and filter $E_{cand}$ in a single pass:
\begin{equation}
\label{eq:gef}
E_{final}=\mathrm{LLM}(E_{cand}, \kappa) .
\end{equation}
Unlike constraint-chain exploration, which executes each constraint through explicit KG queries, GEF relies on the LLM's parametric knowledge to judge constraint satisfaction without additional KG grounding. This makes GEF less faithful but more robust: it can handle constraints that are difficult to formalize as relation chains (e.g., temporal or comparative constraints with no direct KG predicate). Because $E_{cand}$ is still derived from explicit KG execution through the main chain, this layer retains partial KG grounding.

\subsubsection{Layer 2: LLM Direct Generation}

If main-chain reasoning itself fails irrecoverably, the system falls back to CoT-based direct answer generation. The final answer is obtained as
\begin{equation}
E_{final}=\mathrm{LLM}(q),
\end{equation}
where the LLM produces answers conditioned solely on the question without KG grounding. This layer provides no KG-grounded evidence.

Fallback stage and trigger cause are logged for faithfulness analysis.

\section{Faithfulness Metrics}

Answer correctness alone cannot distinguish explicit KG reasoning from parametric memory recall. This section defines two faithfulness metrics to better characterize how well methods ground their answers in external knowledge graphs.

\subsection{Knowledge Grounded Rate (KGR)}

For each evaluation sample, a final grounding-source state $a_i$ is assigned according to the evidence source used to produce the final answer.
Three states are considered, corresponding to decreasing levels of KG grounding:
\begin{equation}
\label{eq:kgr_action_set}
a_i \in \{\texttt{structured},\texttt{synthesis},\texttt{parametric}\}.
\end{equation}

The three states are defined in a method-agnostic manner so that they can be applied to any KG-augmented LLM method.
The \texttt{structured} state indicates that the final answer is derived entirely from explicit KG execution, with no parametric inference involved in the answer-producing step.
The \texttt{synthesis} state indicates that KG execution provides only partial intermediate results (e.g., a candidate set) and does not directly produce final answers; final answer determination requires the LLM to reason over these intermediate results by integrating parametric knowledge.
The \texttt{parametric} state indicates that final answers are produced directly from LLM parametric knowledge without executable KG evidence.
For all KG-augmented LLM methods, outcomes are categorized into these three states under the same criteria and protocol.

Specifically in CCoR, \texttt{structured} denotes that the model completes exploration and obtains the final answer set $E_{final}$ without triggering fallback.
\texttt{synthesis} denotes that exploration falls back to Layer~1 (Global Entity Filtering).
\texttt{parametric} denotes that exploration falls back to Layer~2 (LLM direct generation).

Each state is assigned a weight reflecting its degree of KG grounding:

\begin{equation}
\label{eq:kgr_weights}
w(a_i)=\begin{cases}
1.0, & a_i=\texttt{structured},\\
0.5, & a_i=\texttt{synthesis},\\
0,   & a_i=\texttt{parametric}.
\end{cases}
\end{equation}

For $N$ evaluation samples, KGR is defined as
\begin{equation}
\label{eq:kgr}
\mathrm{KGR}=\frac{1}{N}\sum_{i=1}^{N} w(a_i).
\end{equation}

This metric quantifies how strongly final predictions are supported by explicit KG-grounded reasoning. 
The weighted formulation distinguishes partial grounding in \textit{Fallback Layer~1} from full grounding, so that answers relying on different evidence sources are credited according to their actual degree of KG grounding.

\subsection{Faithful-F1}

Faithful-F1 is motivated by the need to jointly evaluate answer correctness and reasoning faithfulness in a single metric. It combines F1-Score and KGR as follows:
\begin{equation}
\label{eq:faithful_f1}
\mathrm{Faithful\text{-}F1}=\mathrm{KGR}\times \mathrm{F1}.
\end{equation}

This metric penalizes cases where answer correctness is high but KG grounding is weak, and therefore better reflects faithful complex-KGQA performance.

\section{Experiments}

\subsection{Datasets}

This work evaluates CCoR on four KGQA datasets from two knowledge graphs: WebQSP, CWQ, and GrailQA on Freebase\footnote{https://developers.google.com/freebase}, and QALD10-EN on Wikidata\footnote{https://www.wikidata.org/}. 
Table~\ref{tab:dataset_stats} reports statistics on main-chain depth and whether constraints are present per dataset. 

\begin{itemize}
\item \textbf{WebQSP}~\cite{yih2016value} contains 1,639 test questions on Freebase. The majority require 1--2-hop reasoning (1,033 single-hop and 595 two-hop), and most questions are unconstrained (1,232 out of 1,639), making it a relatively straightforward benchmark.

\item \textbf{ComplexWebQuestions (CWQ)}~\cite{talmor2018web} contains 3,520 test questions on Freebase. Two-hop reasoning dominates (1,597), with a substantial portion requiring three or more hops (560). Most questions are constrained (2,266 out of 3,520), posing a significantly harder challenge than WebQSP.

\item \textbf{GrailQA}~\cite{gu2021beyond} contains 1,000 test questions on Freebase. Most require single-hop reasoning (769), yet nearly all are constrained (974 out of 1,000), making it a constraint-rich benchmark designed to evaluate compositional and zero-shot generalization over a broad ontology.

\item \textbf{QALD10-EN}~\cite{perevalov2022qald} contains 333 test questions over Wikidata and emphasizes compositional reasoning with diverse constraint types. Since QALD10-EN does not provide parseable SPARQL annotations, its depth and constraint statistics are unavailable.
\end{itemize}

\subsection{Baselines}
To provide a comprehensive evaluation of CCoR, this paper compares it against a diverse set of strong baselines.

\begin{itemize}
\item \textbf{Tuning-based.} FC-KBQA~\cite{zhang2023fc} decouples knowledge acquisition from reasoning via fine-to-coarse composition. Pangu~\cite{gu2023don} focuses on compositional generalization for complex query structures. FlexKBQA~\cite{li2024flexkbqa} adapts to new KGs with few-shot learning. GAIN~\cite{shu2024distribution} augments training data to improve robustness against distribution shifts. These methods involve task-specific fine-tuning. 

\item \textbf{Retrieval-based.} UniKGQA~\cite{jiangunikgqa} unifies retrieval and reasoning within a single model for multi-hop question answering. DecAF~\cite{yudecaf} jointly decodes answers and logical forms to retrieve relevant KG facts. RoG~\cite{luoreasoning} retrieves faithful reasoning paths from the KG to guide LLM reasoning. 
ELMK~\cite{liao2026enhancing} trains a multi-path encoder to retrieve semantically diverse reasoning graphs as KG evidence.
These methods supply retrieved KG evidence as in-context prompts in an end-to-end manner.

\item \textbf{Agent-based.} As the most basic agents without KG interaction, IO Prompting performs direct question answering without intermediate reasoning steps, and Chain-of-Thought (CoT)~\cite{wei2022chain} prompts the LLM to generate step-by-step reasoning with the instruction ``Let's think step by step.'' Building on KG interaction, StructGPT~\cite{jiang2023structgpt} equips the LLM with specialized programmatic interfaces to interactively query structured data. ToG~\cite{sunthink} employs entity-centric exploration with beam search. PoG~\cite{chen2024plan} uses an adaptive planning framework with a self-correction mechanism. These agent-based methods are training-free.
\end{itemize}

\subsection{Implementation Details}

Experimental setup details are organized into three aspects: model configuration, agent settings, and knowledge graph resources.

\begin{itemize}
\item \textbf{LLM Backbone Configuration.} CCoR and the most closely related baselines (IO Prompting, CoT, ToG, and PoG) are evaluated under three backbone models in the main experiments: GPT-3.5-turbo, GPT-4.1-mini, and DeepSeek-V3.1. GPT-5.4-mini is included in the complex query evaluation to examine CCoR's behavior under a stronger backbone. StructGPT is included only via their published GPT-3.5-turbo results. Tuning-based methods use their own published model checkpoints and are not subject to backbone variation. All models use a temperature of 0.01 for deterministic reasoning.

\item \textbf{Agent-based Methods Configuration.} For agent-based methods, the exploration process uses top-$k$ pruning with $k=3$ as the beam width for selecting relations or entities at each step. The maximum reasoning depth is set to $l=3$ hops on WebQSP, QALD10 and GrailQA, and $l=4$ hops on CWQ.

\item \textbf{Knowledge Graph Setup.} This work uses the official Freebase snapshot following the setup in ToG~\cite{sunthink}. For Wikidata, this work uses the official SPARQL endpoint\footnote{https://query.wikidata.org/sparql}. All methods use the same KG to ensure a fair comparison.
\end{itemize}

\subsection{Evaluation Metrics}

This work reports metrics for both answer correctness and reasoning faithfulness.

\begin{itemize}
\item \textbf{Hits@1.} Measures question-level correctness, i.e., whether at least one predicted answer matches any gold answer entity.

\item \textbf{Precision, Recall, and F1.} Measure answer-set quality by evaluating exactness, completeness, and their balance over predicted and gold entities.

\item \textbf{Knowledge Grounded Rate (KGR).} Measures reasoning faithfulness as a weighted grounding score over final reasoning states, including full grounding, partial grounding, and parametric fallback.

\item \textbf{Faithful-F1 (FF1).} Measures faithfulness-aware answer quality by combining answer correctness with reasoning grounding.
\end{itemize}

\subsection{Main Results}

\begin{table}[t]
\centering
\caption{Dataset statistics by main-chain depth and constraint-chain count.}
\label{tab:dataset_stats}
\small
\begin{tabular}{lcccccc}
\toprule
\multirow{2}{*}{Dataset} & \multirow{2}{*}{Total} & \multicolumn{3}{c}{Depth} & \multicolumn{2}{c}{Constraints} \\ \cmidrule(lr){3-5} \cmidrule(lr){6-7}
 & & 1 & 2 & $\geq$3 & N & Y \\
\midrule
WebQSP    & 1639 & 1033 & 595 & 6 & 1232 & 402 \\
CWQ       & 3520 & 1363 & 1597 & 560 & 1254 & 2266 \\
GrailQA   & 1000 & 769 & 222 & 9 & 26 & 974  \\
QALD10 & 333 & - & - & - & - & - \\
\bottomrule
\end{tabular}
\end{table}

\begin{table*}[t]
\centering
\caption{Accuracy Evaluation. Results of baseline methods are taken from their original papers for comparison; ``--'' indicates that the corresponding result is not reported in the original paper.}
\small
\setlength{\tabcolsep}{3pt}
\begin{tabular}{lcccccccc}
\toprule
\multirow{2}{*}{Method} 
& \multicolumn{2}{c}{WebQSP} 
& \multicolumn{2}{c}{CWQ} 
& \multicolumn{2}{c}{QALD10} 
& \multicolumn{2}{c}{GrailQA} \\ 
\cmidrule(lr){2-3} \cmidrule(lr){4-5} \cmidrule(lr){6-7} \cmidrule(lr){8-9}
& Hits@1 & F1 
& Hits@1 & F1 
& Hits@1 & F1 
& Hits@1 & F1 \\
\midrule

\multicolumn{9}{c}{Tuning-based Methods} \\
\midrule
FC-KBQA 
& -- & 76.9 
& -- & 56.4 
& -- & -- 
& -- & -- \\

Pangu 
& -- & 79.6 
& -- & -- 
& -- & -- 
& -- & 81.7 \\

Pangu (100-shot) w/o tuning 
& -- & 54.5 
& -- & -- 
& -- & -- 
& -- & 62.7 \\

Pangu (10-shot) w/o tuning 
& -- & 45.9 
& -- & -- 
& -- & -- 
& -- & 56.3 \\

FlexKBQA 
& -- & 60.6 
& -- & -- 
& -- & -- 
& -- & 69.4 \\

\midrule
\multicolumn{9}{c}{Retrieval-based Methods} \\
\midrule

UniKGQA 
& 77.2 & 72.2 
& 51.2 & 49.1 
& -- & -- 
& -- & -- \\

DECAF 
& 82.1 & 78.8 
& -- & -- 
& -- & -- 
& -- & -- \\

RoG 
& 85.7 & 70.8 
& 62.6 & 56.2 
& -- & -- 
& -- & -- \\

ELMK 
& 83.6 & 72.5
& 63.0 & 60.6 
& -- & -- 
& -- & -- \\

\midrule
\multicolumn{9}{c}{Agent-based Methods w/ GPT-3.5-turbo} \\
\midrule
IO-Prompt 
& 62.7 & 47.9 
& 35.6 & 31.6 
& 45.0 & 41.6 
& 33.1 & 27.9 \\

CoT-Prompt 
& 60.5 & 45.9 
& 38.6 & 34.9 
& 47.9 & 44.8 
& 33.2 & 28.1 \\

StructGPT 
& 72.6 & -- 
& -- & -- 
& -- & -- 
& -- & -- \\

ToG 
& 76.2 & -- 
& 58.9 & -- 
& 50.2 & -- 
& 68.7 & -- \\

PoG 
& 82.0 & -- 
& 63.2 & -- 
& -- & -- 
& 76.5 & -- \\

CCoR (ours) 
& \textbf{85.1} & \textbf{77.9} 
& \textbf{66.0} & \textbf{57.5}
& \textbf{67.9} & \textbf{64.9} 
& \textbf{79.8} & \textbf{76.1} \\

\bottomrule
\end{tabular}
\label{tab:main-results}
\end{table*}

\subsubsection{Accuracy Evaluation}

Table~\ref{tab:main-results} reports Hits@1 and F1 under GPT-3.5-turbo across the four benchmarks. 
Overall, CCoR achieves the best accuracy on CWQ and QALD10, outperforming all baselines including those that rely on supervised training. On WebQSP and GrailQA, CCoR remains the strongest among training-free methods and is only marginally behind the best supervised-training methods. 
Moreover, these benchmarks cover both Freebase (WebQSP, CWQ, GrailQA) and Wikidata (QALD10), and CCoR performs strongly on both, showing that it transfers well across heterogeneous knowledge graphs.

Compared with tuning-based methods, CCoR reaches comparable accuracy without any training. Pangu is a representative case. With full fine-tuning, it achieves 79.6 F1 on WebQSP and 81.7 on GrailQA, slightly ahead of CCoR (e.g., by 1.7 F1 on WebQSP). However, this advantage relies entirely on task-specific fine-tuning over the target KG. Once that fine-tuning is removed, Pangu collapses on GrailQA, dropping from 81.7 F1 to 62.7 in the 100-shot setting and to 56.3 in the 10-shot setting, far below CCoR. This contrast indicates that the strength of tuning-based methods is tied to retraining on each target KG and does not transfer once such supervision is unavailable. CCoR, by contrast, is training-free and uses at most five in-context examples in its prompts, yet remains competitive across all four benchmarks rather than a single KG-specific subset.

Compared with retrieval-based methods, the comparison concentrates on WebQSP and CWQ, the two Freebase benchmarks on which these methods report results. On the simpler WebQSP, CCoR is only marginally behind the strongest retrieval-based methods, which is expected since their retrievers are supervised-trained while CCoR is training-free. 
On the harder CWQ, however, CCoR surpasses the best retrieval-based method RoG, improving Hits@1 by 3.4 and F1 by 1.3. This contrast shows that CCoR matches strong retrieval-based methods on simple queries and becomes more effective as query complexity grows.

Compared within the agent-based group, all methods are training-free, and CCoR attains the best Hits@1 and F1 across all four benchmarks. 
The comparison further reveals two sources of gain. First, IO-Prompt and CoT-Prompt are simple prompt-augmented agents that answer without KG access and perform worst. Coupling the LLM with the KG brings a large jump, confirming that grounding reasoning in the KG is essential. 
Second, the relation-centric CCoR is consistently stronger than the entity-centric ToG and PoG, which shows the benefit of shifting the search unit from entities to relations together with grounding constraint handling in the KG. This advantage is largest on the constraint-heavy CWQ, where CCoR improves Hits@1 from 63.2 (PoG) to 66.0, while it stays marginal on the lightly constrained WebQSP. This pattern indicates that the gain of CCoR grows precisely when queries involve explicit constraints, where explicit KG execution replaces unreliable parametric judgment.

\begin{table}[ht]
\centering
\caption{Faithfulness Evaluation on Agent-based Methods. ($^\dagger$: reproduced from official codebases).}
\small
\setlength{\tabcolsep}{5pt}
\begin{tabular}{lcccccc}
\toprule
\multirow{2}{*}{Method}
& \multicolumn{3}{c}{WebQSP}
& \multicolumn{3}{c}{CWQ} \\
\cmidrule(lr){2-4} \cmidrule(lr){5-7}
& F1 & KGR & FF1
& F1 & KGR & FF1 \\
\midrule

\multicolumn{7}{c}{GPT-4.1-mini} \\
\midrule
IO-Prompt
& 47.7 & 0.0 & 0.0
& 32.1 & 0.0 & 0.0 \\

CoT-Prompt
& 46.4 & 0.0 & 0.0
& 35.2 & 0.0 & 0.0 \\

ToG$^\dagger$
& 57.5 & 56.5 & 32.5
& 41.2 & 30.3 & 12.5 \\

PoG$^\dagger$
& 64.0 & 63.3 & 40.5
& 42.2 & 37.1 & 15.7 \\

CCoR (ours)
& \textbf{78.0} & \textbf{90.4} & \textbf{70.5}
& \textbf{57.8} & \textbf{69.1} & \textbf{39.9} \\

\midrule
\multicolumn{7}{c}{DeepSeek-V3.1} \\
\midrule
ToG$^\dagger$
& 56.3 & 56.4 & 31.8
& 37.3 & 28.6 & 10.7 \\

PoG$^\dagger$
& 64.6 & 62.8 & 40.6
& 41.4 & 35.8 & 14.8 \\

CCoR (ours)
& \textbf{73.5} & \textbf{86.1} & \textbf{63.3}
& \textbf{60.3} & \textbf{74.2} & \textbf{44.7} \\

\bottomrule
\end{tabular}
\label{tab:faithfulness-evaluation}
\end{table}

\subsubsection{Faithfulness Evaluation on Agent-based Methods}

Table~\ref{tab:faithfulness-evaluation} reports F1, KGR, and FF1 on WebQSP and CWQ under GPT-4.1-mini and DeepSeek-V3.1. Unlike F1, which only measures answer correctness, KGR and FF1 reward answers that are actually grounded in KG execution, so they expose how each method reaches its answers. The results separate the agent-based methods into three tiers.
First, IO-Prompt and CoT-Prompt answer purely from parametric knowledge without any KG access. They still reach non-trivial F1, for example 47.7 and 46.4 on WebQSP under GPT-4.1-mini, but their KGR and FF1 are both 0.0 on both datasets. This shows that answer correctness alone can fully hide the absence of grounding, which is exactly what the faithfulness metrics are designed to reveal.
Second, ToG and PoG add KG interaction under entity-centric exploration and lift KGR above zero, yet their grounding stays limited. On CWQ under GPT-4.1-mini, their KGR is only 30.3 and 37.1, so a large share of answers still falls back on parametric knowledge, which keeps FF1 low at 12.5 and 15.7. The same pattern holds under DeepSeek-V3.1, indicating that entity-centric exploration grounds only part of its reasoning regardless of the backbone.
Third, the relation-centric CCoR improves all three metrics by a clear margin over the entity-centric methods, and the gain is most visible on KGR. Under GPT-4.1-mini, CCoR lifts KGR from 63.3 (PoG) to 90.4 on WebQSP and from 37.1 to 69.1 on CWQ, raising FF1 from 40.5 to 70.5 and from 15.7 to 39.9. The same advantage holds under DeepSeek-V3.1, where CCoR achieves KGR scores of 86.1 and 74.2 on the two datasets respectively, far above the entity-centric baselines. 
This sharp KGR gain stems from the constraint chain, which resolves constraints through explicit KG execution rather than parametric judgment and thus keeps far more answers grounded in the KG. The stronger grounding further translates into higher F1, where CCoR remains the best on both datasets and backbones. 
This confirms that faithfulness and accuracy improve together rather than trading off.

\begin{table}[ht]
\centering
\caption{Results on CWQ Complex Query ($^\dagger$: reproduced from official codebases).}
\label{tab:cwq_setting}
\small
\setlength{\tabcolsep}{4pt}
\begin{tabular}{lcccccc}
\toprule
\multirow{2}{*}{Method} & \multicolumn{6}{c}{Complex Query (500)} \\
\cmidrule(lr){2-7}
 & Hits@1 & P & R & F1 & KGR & FF1 \\
\midrule
\multicolumn{7}{c}{GPT-4.1-mini} \\
\midrule
ToG$^\dagger$          & 50.8 & 50.3 & 48.6 & 48.8 & 24.2 & 11.8 \\
PoG$^\dagger$          & 54.2 & 53.0 & 51.9 & 51.6 & 36.4 & 18.8 \\
CCoR         & \textbf{68.4} & \textbf{61.6} & \textbf{67.3} & \textbf{62.3} & \textbf{61.9} & \textbf{38.6} \\
\midrule
\multicolumn{7}{c}{GPT-5.4-mini} \\
\midrule
CCoR (ours)        & \textbf{70.2} & \textbf{64.3} & \textbf{69.2} & \textbf{65.1} & \textbf{77.3} & \textbf{50.3} \\
\bottomrule
\end{tabular}
\end{table}

\subsection{Performance on Complex Query}

In this subsection, ``complex queries'' refer to CWQ questions that contain explicit constraints, i.e., samples with constraint-chain count $\geq 1$. To build a focused evaluation set, this work filters the CWQ test split with this criterion and then uniformly samples 500 questions at random. The purpose is to directly test CCoR in constraint scenarios, where its key design claim is to improve both answer accuracy and reasoning faithfulness through explicit constraint-chain execution.

Table~\ref{tab:cwq_setting} confirms that CCoR's gains center on exactly these two aspects. CCoR achieves the best Hits@1 and FF1 under both backbones, consistently outperforming all baselines by a wide margin. Under GPT-4.1-mini, CCoR lifts Hits@1 from 54.2 (PoG) to 68.4 and FF1 from 18.8 to 38.6, with the largest separation appearing on KGR (36.4 to 61.9). This shows that on constraint-heavy queries, the advantage of CCoR comes mainly from grounding constraint verification in explicit KG execution rather than parametric judgment. 
The advantage further strengthens with a stronger backbone: moving from GPT-4.1-mini to GPT-5.4-mini, CCoR improves consistently across all metrics, reaching a Hits@1 of 70.2 and an FF1 of 50.3. 
Because constraint verification is grounded in KG execution, CCoR decouples reasoning quality from the variability of parametric memory across models, so a stronger backbone translates directly into more faithful and more accurate constraint handling.

\subsection{Efficiency Study}

\begin{figure*}[t]
\centering
\includegraphics[width=0.8\textwidth]{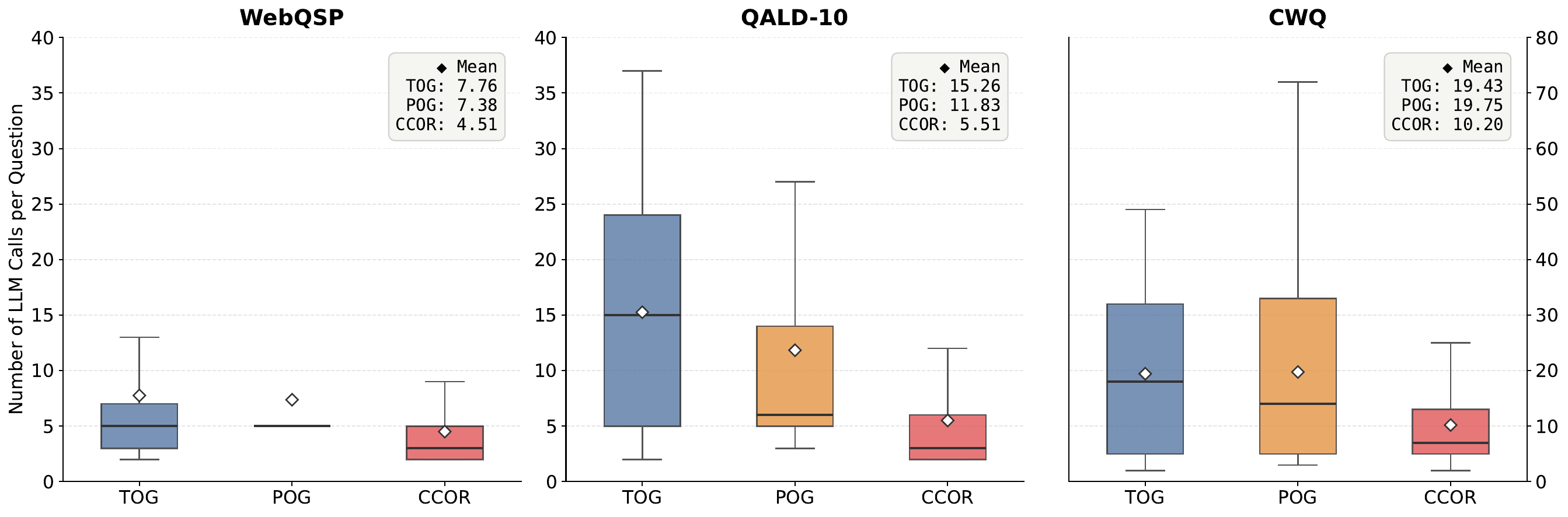}
\caption{Distribution of LLM API calls per question for CCoR, ToG, and PoG on WebQSP, CWQ, and QALD10. CCoR consistently requires fewer calls than ToG and PoG.}
\label{fig:call}
\end{figure*}

\subsubsection{LLM Call Distribution}

Figure~\ref{fig:call} compares the distribution of LLM API calls per question for CCoR, ToG, and PoG across three datasets with GPT-4.1-mini. 
On average, CCoR consistently issues fewer calls across all three datasets. On WebQSP, it uses only 4.51 calls, roughly half of the 7.76 for ToG$^\dagger$ and 7.38 for PoG$^\dagger$. On QALD10, it needs 5.51 calls, far below the 15.26 for ToG$^\dagger$ and 11.83 for PoG$^\dagger$. On CWQ, it needs 10.20 calls, about half of the 19.43 for ToG$^\dagger$ and 19.75 for PoG$^\dagger$, even though it additionally runs constraint-chain exploration. 
The savings on CWQ come from explicit constraint-chain execution, which separates constraint verification from main-chain exploration and thus avoids the redundant branch expansion and backtracking that entity-centric baselines spend on constraint handling. 
The medians and interquartile ranges show the same trend, confirming that relation-centric exploration resolves most questions with consistently fewer LLM calls.

\begin{table*}[t]
\centering
\caption{Case study on the question ``What team with fight song Bow Down to Washington did Nate Robinson play for?'' (Gold answer: \{Washington Huskies football\}).}

\label{tab:case_study}
\small
\begin{tabular}{lp{2.8cm}p{7.5cm}crr}
\toprule
Method & Predicted Answer & Reasoning Process & Calls & Hits@1 & F1 \\
\midrule
CoT & 
University of Washington & 
Parametric recall only; identifies the university but does not distinguish specific sports teams. & 
1 & 
0 & 0 \\
\midrule
ToG & 
University of Washington & 
Entity-centric exploration fails to reach the team-level granularity, falls back to parametric generation. & 
32 & 
0 & 0 \\
\midrule
PoG & 
University of Washington & 
Entity-centric exploration with entity-level backtracking, but repeatedly revisits incorrect paths without resolving the fight-song constraint, falls back to parametric generation. & 
30 & 
0 & 0 \\
\midrule
CCoR (ours) & 
\textbf{Washington Huskies football} & 
Main chain: Nate Robinson $\xrightarrow{\texttt{pro\_athlete.teams}}$ $e_1$ $\xrightarrow{\texttt{team\_roster.team}}$ 8 candidate teams. \newline
Constraint chain: $?$team $\xrightarrow{\texttt{fight\_song}}$ Bow Down to Washington, filters candidates to \{Washington Huskies football\}. & 
\textbf{6} & 
\textbf{1} & \textbf{1.00} \\
\bottomrule
\end{tabular}
\end{table*}

\subsubsection{Token Consumption}

\begin{table}[htbp]
\centering
\caption{Total token consumption on GPT-4.1-mini ($^\dagger$: reproduced from official codebases).}
\small
\begin{tabular}{lccc}
\toprule
Method & WebQSP & QALD10 & CWQ \\
\midrule
ToG$^\dagger$ & 11,591,404 & 3,075,735 & 62,348,816 \\
PoG$^\dagger$ & \textbf{11,340,446} & 5,713,781 & 113,186,196 \\
CCoR (ours) & 12,202,046 & \textbf{2,493,586} & \textbf{59,172,341} \\
\bottomrule
\end{tabular}
\label{tab:efficiency_totals}
\end{table}

Table~\ref{tab:efficiency_totals} reports total token consumption with GPT-4.1-mini. 
CCoR is more efficient on complex queries while remaining comparable on simple ones. 
On the complex CWQ and QALD10, accurate constraint-chain exploration grounds each constraint with explicit KG execution and thereby avoids the redundant entity-centric retrieval and backtracking that PoG spends on constraint handling, so CCoR consumes the fewest tokens, about 59.2M on CWQ and 2.5M on QALD10, with the gap widening to roughly half of PoG's 113.2M on CWQ. 
On the simple WebQSP, by contrast, few queries trigger the constraint chain, so its extra exploration adds only a small cost, and CCoR uses slightly more tokens than PoG, though the gap stays within one million. 
Overall, the constraint chain costs little on simple queries but yields large savings on complex ones, and the modest overhead on WebQSP is outweighed by the substantial gains in accuracy and faithfulness reported in the main results.

\subsection{Impact of Reasoning Depth}

\begin{table}[ht]
\centering
\caption{Effect of reasoning depth on CWQ (GPT-4.1-mini).}
\begin{tabular}{lcccc}
\toprule
Depth & Cum. Ratio (\%) & Hits@1 & F1 & KGR \\
\midrule
2 & 84.1 & 47.8 & 43.7 & 50.3 \\
3 & 96.0 & 62.8 & 50.6 & 63.0 \\
4 & 100.0 & \textbf{67.0} & \textbf{57.8} & \textbf{69.1} \\
5 & 100.0 & 64.7 & 53.9 & 68.2 \\
\bottomrule
\end{tabular}
\label{tab:cwq_depth}
\end{table}

Table~\ref{tab:cwq_depth} evaluates CCoR under varying maximum reasoning depths on CWQ with GPT-4.1-mini. Here, Cum. Ratio denotes the proportion of questions whose required reasoning depth is no greater than the configured maximum depth. As depth increases from 2 to 4, Cum. Ratio rises from 84.1\% to 100.0\%, and Hits@1/F1/KGR improve from 47.8/43.7/50.3 to 67.0/57.8/69.1. This indicates that depth 4 provides the best trade-off between exploration completeness and error accumulation. Although depth 5 keeps full coverage (100.0\%), performance declines to 64.7 Hits@1, 53.9 F1, and 68.2 KGR, suggesting that deeper exploration introduces noisy branches without expanding the reachable question set.

\subsection{Ablation Study}

\begin{table}[ht]
\centering
\caption{Ablation Study on CWQ using GPT-4.1-mini.}
\begin{tabular}{lccc}
\toprule
 & Hits@1 & F1 & KGR \\
\midrule
CCoR			 & \textbf{67.0}   & \textbf{57.8} & \textbf{69.1} \\
\quad w/o Constraint Chain (CoR)  & 60.3  & 52.0 & 49.8 \\ 
\quad w/o Relation Backtrack       & 57.2   & 50.0 & 42.6 \\
\quad w/o Global Entity Filtering      & 48.2   & 42.8 & 29.8 \\ 
\bottomrule
\end{tabular}
\label{tab:ablation_study}
\end{table}

Table~\ref{tab:ablation_study} isolates the contribution of each CCoR component on CWQ with GPT-4.1-mini by cumulatively removing modules, where each row drops one more component on top of the previous one. 
The first removal is constraint-chain exploration. The pipeline then keeps only main-chain exploration and resolves constraints through Global Entity Filtering, replacing explicit KG execution with implicit judgment, which is equivalent to CoR.
This causes a clear drop across all three metrics, with the sharpest fall on KGR, from 69.1 to 49.8. 
The drop shows that constraint-chain exploration improves not only final answers but also KG-grounded execution quality. 
This matches the design goal of CCoR: constraints are resolved through explicit KG operations rather than implicit judgments.

The second removal is relation-level backtracking. The pipeline then keeps a single main-chain branch and cannot recover from early relation-selection errors during multi-hop traversal. This causes only a milder decline, with KGR slipping to 42.6. The mild drop shows that backtracking mainly serves as a robustness mechanism for upstream chain construction rather than a core source of grounding.

The third removal is Global Entity Filtering, which leaves only main-chain exploration with no constraint handling at all. This causes the largest degradation, with KGR dropping sharply to 29.8. The sharp drop shows that constraint handling is the key bottleneck on CWQ, where constrained questions are prevalent. 
Overall, the three removals show a clear hierarchy of contributions. Constraint handling is the dominant source of both accuracy and grounding, while relation-level backtracking acts as a secondary robustness mechanism for upstream chain construction. This hierarchy explains why CCoR delivers the largest gain in faithfulness by replacing implicit constraint filtering with explicit constraint-chain execution.

\subsection{Case Study}

Table~\ref{tab:case_study} compares four methods on a constrained two-hop question from CWQ. The question requires first reaching the team level through multi-hop traversal and then enforcing a lexical constraint (fight song) on the final team entity. This combination is challenging because many methods can retrieve the university context but fail to execute the final team-level constraint precisely.

CoT, ToG, and PoG all fail on this case. CoT directly relies on parametric recall and outputs ``University of Washington.'' ToG and PoG spend many more calls but still converge to the same university-level answer, indicating that additional exploration alone does not resolve the constraint when the reasoning process remains entity-centric and does not execute the fight-song condition explicitly. As a result, all three methods obtain zero Hits@1/F1.

CCoR succeeds by grounding both phases in the KG. Its main chain builds the two-hop relation path \texttt{pro\_athlete.teams} $\rightarrow$ \texttt{team\_roster.team} and retrieves 8 candidate teams that include the gold answer, correctly reaching the team-level granularity that the entity-centric methods miss. The remaining difficulty is to single out the team whose fight song matches the constraint. Instead of judging this implicitly, CCoR executes an explicit constraint chain, $?$team $\xrightarrow{\texttt{fight\_song}}$ Bow Down to Washington, which filters the 8 candidates over the KG and returns only \{Washington Huskies football\}, yielding Hits@1 of 1 and F1 of 1.00. The total cost is only 6 calls, far fewer than the 30--32 spent by ToG and PoG. 
This case therefore provides direct evidence that explicit constraint-chain execution is the key mechanism behind CCoR's accuracy and faithfulness improvement.

\section{Conclusion}

This paper presents Compositional Chain-of-Relations (CCoR), an agent-based framework for faithful complex KGQA. Built on the relation-centric exploration paradigm, CCoR composes two relation chains to mirror the two phases of complex query solving: a main chain for candidate retrieval and an on-demand constraint chain for explicit constraint execution over the KG.
Experiments on four KGQA benchmarks and multiple LLM backbones show that CCoR consistently improves accuracy, faithfulness, and efficiency over strong baselines, with the most pronounced gains on constraint-heavy queries. Among these three aspects, the faithfulness improvement is the most distinctive. This advantage is reflected by substantial gains on the proposed Knowledge Grounded Rate (KGR) and Faithful-F1 (FF1), indicating that CCoR more consistently grounds its answers in explicit KG reasoning.

Although relation-centric exploration requires fewer LLM calls than entity-centric methods by avoiding entity pruning, CoR/CCoR still requires multiple LLM calls in complex KG reasoning. At the current stage, this cost is a necessary trade-off for maintaining faithfulness through explicit KG grounding. Reducing call cost while preserving faithfulness remains a key direction for future work.

\bibliographystyle{IEEEtran}
\bibliography{ref}


\end{document}